\documentclass[11pt,a4paper]{article}
\usepackage[hyperref]{acl2018}
\usepackage{times}
\usepackage{latexsym}
\usepackage{amsmath}
\usepackage{amssymb}
\usepackage{comment}
\usepackage{bbm}
\usepackage{multirow}
\usepackage{enumitem}
\usepackage[normalem]{ulem}
\usepackage{inconsolata}

\usepackage{url}

\usepackage{graphicx}
\usepackage{xcolor}

\newcommand{\eg}{\textit{e.g.} }

\aclfinalcopy 

\title{Extreme Adaptation for Personalized Neural Machine Translation}

\author{Paul Michel \\
  Language Technologies Institute \\
  Carnegie Mellon University \\
  {\tt pmichel1@cs.cmu.edu} \\\And
  Graham Neubig \\
  Language Technologies Institute \\
  Carnegie Mellon University \\
  {\tt gneubig@cs.cmu.edu} \\}

\date{}

\begin{document}
\maketitle
\begin{abstract}
  Every person speaks or writes their own flavor of their native language, influenced by a number of factors: the content they tend to talk about, their gender, their social status, or their geographical origin.
  When attempting to perform Machine Translation (MT), these variations have a significant effect on how the system should perform translation, but this is not captured well by standard one-size-fits-all models.
  In this paper, we propose a simple and parameter-efficient adaptation technique that only requires adapting the bias of the output softmax to each particular user of the MT system, either directly or through a factored approximation.
  Experiments on TED talks in three languages demonstrate improvements in translation accuracy, and better reflection of speaker traits in the target text.
\end{abstract}

\section{Introduction}

The production of language varies depending on the speaker or author, be it to reflect personal traits (\eg job, gender, role, dialect) or the topics that tend to be discussed (\eg technology, law, religion). Current Neural Machine Translation (NMT) systems do not incorporate any explicit information about the speaker, and this forces the model to learn these traits implicitly. This is a difficult and indirect way to capture inter-personal variations, and in some cases it is impossible without external context (Table 1, \newcite{mirkin-EtAl:2015:EMNLP}).

\begin{table}[!tb]
\small
\centering
\begin{tabular}{ll}

 Source                                   & Translation \\ \hline
 \multirow{2}{*}{I went home}   & {\small\color{gray}[Man]:} Je suis \textcolor{red}{rentr\'e} \`{a} la maison \\
  & {\small\color{gray}[Woman]:} Je suis \textcolor{red}{rentr\'ee} \`{a} la maison \\
\hline
\multirow{2}{*}{I do drug testing}   & {\small\color{gray}[Doctor]:} Je \textcolor{red}{teste des m\'{e}dicaments}\\
& {\small\color{gray}[Police]:} Je \textcolor{red}{d\'{e}piste des drogues}\\\hline
\end{tabular}
\caption{\label{tab:perso_example}Examples where speaker information influences English-French translation. }
\end{table}

Recent work has incorporated side information about the author such as personality \cite{mirkin-EtAl:2015:EMNLP}, gender \cite{rabinovich-EtAl:2017:EACLlong} or politeness \cite{sennrich-haddow-birch:2016:N16-1}, but these methods can only handle phenomena where there are explicit labels for the traits.
Our work investigates how we can efficiently model speaker-related variations to improve NMT models.

In particular, we are interested in improving our NMT system given few training examples for any particular speaker. We propose to approach this task as a domain adaptation problem with an extremely large number of domains and little data for each domain, a setting where we may expect traditional approaches to domain adaptation that adjust all model parameters to be sub-optimal (\S\ref{sec:def}).
Our proposed solution involves modeling the speaker-specific variations as an additional bias vector in the softmax layer, where we either learn this bias directly, or through a factored model that treats each user as a mixture of a few prototypical bias vectors~(\S\ref{sec:model}).

We construct a new dataset of Speaker Annotated TED talks (SATED, \S\ref{sec:dataset}) to validate our approach.
Adaptation experiments (\S\ref{sec:exp_known}) show that explicitly incorporating speaker information into the model improves translation quality and accuracy with respect to speaker traits.%
\footnote{Data/code publicly available at \url{http://www.cs.cmu.edu/~pmichel1/sated/} and \url{https://github.com/neulab/extreme-adaptation-for-personalized-translation} respectively.}

\section{Problem Formulation and Baselines}
\label{sec:def}

In the rest of this paper, we refer to the person producing the source sentence (speaker, author, etc\ldots) generically as the \emph{speaker}. We denote as $\mathcal{S}$ the set of all speakers.

The usual objective of NMT is to find parameters $\theta$ of the conditional distribution $p(y\mid x;\theta)$ to maximize the empirical likelihood.
We argue that personal variations in language warrant decomposing the empirical distribution into $\vert \mathcal{S}\vert$ speaker specific domains $\mathcal{D}_s$ and learning a different set of parameters $\theta_s$ for each.
This setting exhibits specific traits that set it apart from common domain adaptation settings:

\begin{enumerate}[itemsep=0pt]
\item The number of speakers is very large. Our particular setting deals with $\vert \mathcal{S}\vert\approx 1800$ but our approaches should be able to accommodate orders of magnitude more speakers.
\item There is very little data (even monolingual, let alone bilingual or parallel) for each speaker, compared to millions of sentences usually used in NMT.
\item As a consequence of 1, we can assume that many speakers share similar characteristics such as gender, social status, and as such may have similar associated domains.\footnote{Note that the speakers are still unique, and many might use very specific words (\eg the name of their company or of a specific medical procedure that they are an expert on).}
\end{enumerate}


\subsection{Baseline NMT model}
\label{sec:base}

All of our experiments are based on a standard neural sequence to sequence model. We use one layer LSTMs as the encoder and decoder and the \textit{concat} attention mechanism described in \citet{luong2015stanford}. We share the parameters in the embedding and softmax matrix of the decoder as proposed in \citet{press-wolf:2017:EACLshort}. All the layers have dimension 512 except for the attention layer (dimension 256). 
To make our baseline competitive, we apply several regularization techniques such as dropout \cite{srivastava2014dropout} in the output layer and within the LSTM (using the variant presented in \citealp{gal2016theoretically}). We also drop words in the target sentence with probability 0.1 according to \newcite{iyyer-EtAl:2015:ACL-IJCNLP} and implement label smoothing as proposed in \newcite{szegedy2016rethinking} with coefficient $0.1$.
Appendix \ref{sec:model_details} provides a more thorough description of the baseline model.

\subsection{Baseline adaptation strategy}
\label{sec:baseline_adaptation}

As mentioned in \S\ref{sec:def}, our goal is to learn a separate conditional distribution $p(y\mid x, s)$ and parametrization $\theta_s$ to improve translation for speaker $s$. The usual way of adapting from general domain parameters $\theta$ to $\theta_s$ is to retrain the full model on the domain specific data \cite{luong2015stanford}.
Naively applying this approach in the context of personalizing a model for each speaker however has two main drawbacks:

\paragraph{Parameter cost}Maintaining a set of model parameters for each speaker is expensive. For example, the model in \S\ref{sec:base} has $\approx$47M parameters when the vocabulary size is 40k, as is the case in our experiments in \S\ref{sec:exp_known}. Assuming each parameter is stored as a 32bit float, every speaker-specific model costs $\approx$188MB. In a production environment with thousands to billions of speakers, this is impractical.
\paragraph{Overfitting}Training each speaker model with very little data is a challenge, necessitating careful and heavy regularization \cite{micelibarone-EtAl:2017:EMNLP2017} and an early stopping procedure.

\subsection{Domain Token}
\label{sec:domain_token}

A more efficient domain adaptation technique is the \emph{domain token} idea used in \citet{sennrich-haddow-birch:2016:N16-1,chu-dabre-kurohashi:2017:Short}: introduce an additional token marking the domain in the source and/or the target sentence. In experiments, we add a token indicating the speaker at the start of the target sentence for each speaker. We refer to this method as the \texttt{spk\_token} method in the following.

Note that in this case there is now only an embedding vector (of dimension 512 in our experiments) for each speaker. However, the resulting domain embedding are non-trivial to interpret (i.e. it is not clear what they tell us about the domain or speaker itself).

\section{Speaker-specific Vocabulary Bias}
\label{sec:model}

In NMT models, the final choice of which word to use in the next step $t$ of translation is generally performed by the following softmax equation
\begin{equation} \label{eq:bias_dist}
p_t=\text{softmax}(E_To_t + b_T)
\end{equation}
where $o_t$ is predicted in a context-sensitive manner by the NMT system and $E_T$ and $b_T$ are the weight matrix and bias vector parameters respectively.
Importantly, $b_T$ governs the overall likelihood that the NMT model will choose particular vocabulary.
In this section, we describe our proposed methods for making this bias term speaker-specific, which provides an efficient way to allow for speaker-specific vocabulary choice.%
\footnote{Notably, while this limits the model to only handling word choice and does not explicitly allow it to model syntactic variations, favoring certain words over others can indirectly favor certain phenomena (\eg favoring passive speech by increasing the probability of auxiliaries).}

\subsection{Full speaker bias}
\label{sec:full_bias}

We first propose to learn speaker-specific parameters for the bias term in the output softmax only.
This means changing Eq.~\ref{eq:bias_dist} to
\begin{equation} \label{eq:full_bias_dist}
p_t=\text{softmax}(E_To_t + b_T + b_s)
\end{equation}
for speaker $s$. 
This only requires learning and storing a vector equal to the size of the vocabulary, which is a mere 0.09\% of the parameters in the full model in our experiments.
In effect, this greatly reducing the parameter cost and concerns of overfitting cited in \S\ref{sec:baseline_adaptation}. This model is also easy to interpret as each coordinate of the bias vector corresponds to a log-probability on the target vocabulary.
We refer to this variant as \texttt{full\_bias}.

\begin{figure}[!t]
\centering
\includegraphics[width=\columnwidth]{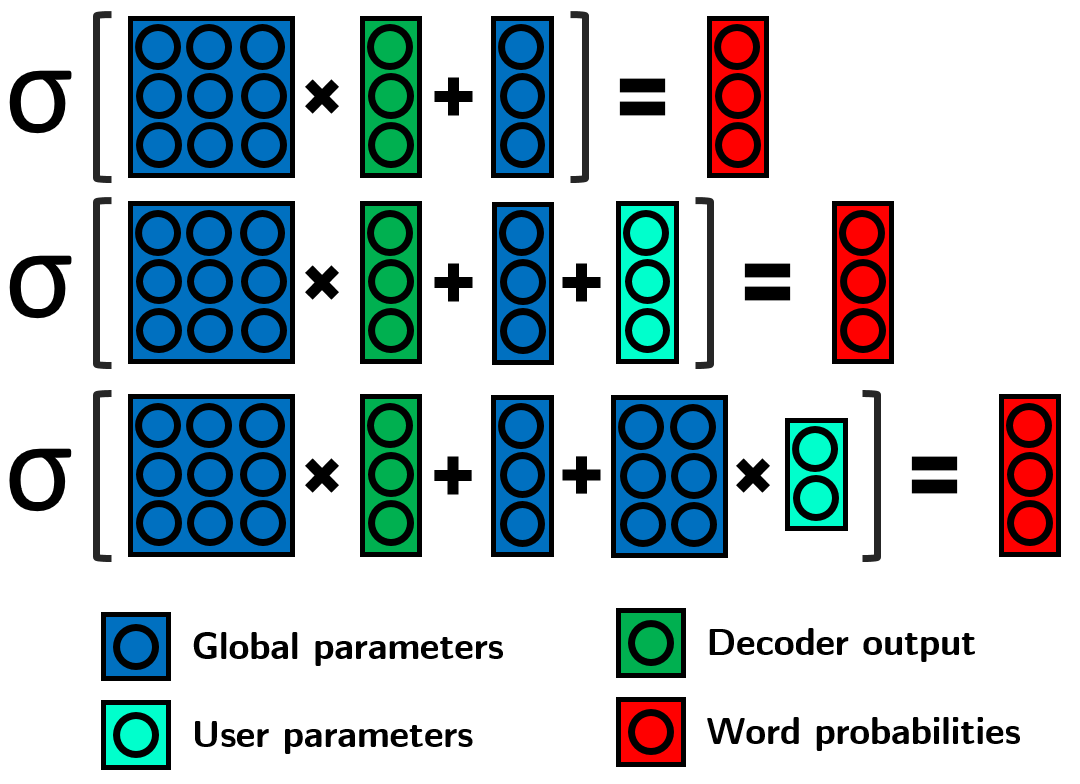}
\caption{\label{fig:models_diagrams}Graphical representation of our different adaptation models for the softmax layer. From top to bottom is the base softmax, the \texttt{full\_bias} softmax and the \texttt{fact\_bias} softmax}
\end{figure}

\subsection{Factored speaker bias}
\label{sec:factored_bias}

The biases for a set of speakers $\mathcal{S}$ on a vocabulary $\mathcal{V}$ can be represented as a matrix:
\begin{equation}
B\in\mathbb R^{\vert\mathcal{S}\vert\times\vert\mathcal{V}\vert}
\end{equation}

\noindent where each row of $B$ is one speaker bias $b_s$.
In this formulation, the $\vert\mathcal{S}\vert$ rows are still linearly independent, meaning that $B$ is high rank. In practical terms, this means that we cannot share information among users about how their vocabulary selection co-varies, which is likely sub-ideal given that speakers share common characteristics.

Thus, we propose another parametrization of the speaker bias, \texttt{fact\_bias}, where the $B$ matrix is factored according to:
\begin{equation}
\begin{split}
B=&S\tilde{B}\\\ S\in&\mathbb R^{\vert\mathcal{S}\vert\times r},\\ \tilde{B}\in&\mathbb R^{r\times\vert\mathcal{V}\vert}
\end{split}
\end{equation}
where $S$ is a matrix of speaker vectors of low dimension $r$ and $\tilde B$ is a matrix of $r$ speaker independent biases.
Here, the bias for each speaker is a mixture of $r$ ``centroid'' biases $\tilde B$ with $r$ speaker ``weights''.
This reduces the total number of parameters allocated to speaker adaptation from $\vert\mathcal{S}\vert\vert\mathcal{V}\vert$ to $r(\vert\mathcal{S}\vert+\vert\mathcal{V}\vert)$. In our experiments, this corresponds to using between $99.38$ and $99.45\%$ fewer parameters than the \texttt{full\_bias} model depending on the language pair, with $r$ parameters per speaker. In this work, we will use $r=10$.

We provide a graphical summary of our proposed approaches in figure \ref{fig:models_diagrams}.

\section{Speaker Annotated TED Talks Dataset}
\label{sec:dataset}

In order to evaluate the effectiveness of our proposed methods, we construct a new dataset, Speaker Annotated TED (SATED) based on TED talks,\footnote{\url{https://www.ted.com}} with three language pairs, English-French (\texttt{en-fr}), English-German (\texttt{en-de}) and English-Spanish (\texttt{en-es}) and speaker annotation.%

The dataset consists of transcripts directly collected from \url{https://www.ted.com/talks}, and contains roughly 271K sentences in each language distributed among 2324 talks. We pre-process the data by removing sentences that don't have any translation or are longer than 60 words, lowercasing, and tokenizing (using the Moses tokenizer \cite{Koehn:2007:MOS:1557769.1557821}).

\begin{table}[tb]
\centering
\begin{tabular}{lcccc}
& \texttt{en-fr} & \texttt{en-es} & \texttt{en-de} \\ \hline
\#talks & 1,887 & 1,922 & 1,670\\
\#train & 177,743 & 182,582 & 156,134\\
\#dev & 3,774 & 3,844 & 3,340\\
\#test & 3,774 & 3,844 & 3,340\\
avg. sent/talk &94,2&95.0&93,5\\
std dev&57,6&57.8&60,3\\
\hline
\end{tabular}
\caption{\label{tab:data_stats}Dataset statistics}
\end{table}

Some talks are partially or not translated in some of the languages (in particular there are fewer translations in German than in French or Spanish), we therefore remove any talk with less than 10 translated sentences in each language pair.

The data is then partitioned into training, validation and test sets. We split the corpus such that the test and validation split each contain 2 sentence pairs from each talk, thus ensuring that all talks are present in every split. Each sentence pair is annotated with the name of the talk and the speaker. Table \ref{tab:data_stats} lists statistics on the three language pairs.

This data is made available under the Creative Commons license, Attribution-Non Commercial-No Derivatives (or the CC BY-NC-ND 4.0 International, \url{https://creativecommons.org/licenses/by-nc-nd/4.0/legalcode}), all credit for the content goes to the TED organization and the respective authors of the talks. The data itself can be found at \url{http://www.cs.cmu.edu/~pmichel1/sated/}.

\section{Experiments}
\label{sec:exp_known}

We run a set of experiments to validate the ability of our proposed approach to model speaker-induced variations in translation.

\subsection{Experimental setup}
\label{sec:exp_known_setup}

We test three models \texttt{base} (a baseline ignoring speaker labels), \texttt{full\_bias} and \texttt{fact\_bias}.
During training, we limit our vocabulary to the 40,000 most frequent words. Additionally, we discard any word appearing less than 2 times. Any word that doesn't satisfy those conditions is replaced with an \texttt{UNK} token.%
\footnote{Recent NMT systems also commonly use sub-word units \cite{sennrich-haddow-birch:2016:P16-12}. This may influence on the result, either negatively (less direct control over high-frequency words) or positively (more capacity to adapt to high-frequency words). We leave a careful examination of these effects for future work.}

All our models are implemented with the DyNet \cite{neubig2017dynet} framework, and unless specified we use the default settings therein. We refer to appendix \ref{sec:training} for a detailed explanation of the training process. We translate the test set using beam search with beam size 5.

\subsection{Does explicitly modeling speaker-related variation improve translation quality?}

Table \ref{tab:known_results} shows final test scores for each model with statistical significance measured with paired bootstrap resampling \cite{koehn:2004:EMNLP}.
As shown in the table, both proposed methods give significant improvements in BLEU score, with the biggest gains in English to French ($+0.99$) and smaller gains in German and Spanish ($+0.74$ and $+0.40$ respectively). Reducing the number of parameters with \texttt{fact\_bias} gives slightly better (\texttt{en-fr}) or worse (\texttt{en-de}) BLEU score, but in those cases the results are still significantly better than the baseline.

\begin{table}[tb]
\centering
\begin{tabular}{lccc}
& en-fr & en-es & en-de \\ \hline
\texttt{base} & 38.05 & 39.89 & 26.46\\ \hline
\texttt{spk\_token} & \textbf{38.85} & 40.04 & 26.52\\ \hline
\texttt{full\_bias} & \textbf{38.54} & \textbf{40.30} & \textbf{27.20}\\ \hline
\texttt{fact\_bias} & \textbf{39.01} & 39.88 & \textbf{26.94}\\ \hline
\end{tabular}
\caption{\label{tab:known_results}Test BLEU. Scores significantly ($p<0.05$) better than the baseline are written in bold}
\end{table}

However, BLEU is not a perfect evaluation metric. In particular, we are interested in evaluating how much of the personal traits of each speaker our models capture. To gain more insight into this aspect of the MT results, we devise a simple experiment.
For every language pair, we train a classifier (continuous bag-of-n-grams; details in Appendix \ref{sec:classifier}) to predict the author of each sentence on the target language part of the training set.
We then evaluate the classifier on the ground truth and the outputs from our 3 models (\texttt{base}, \texttt{full\_bias} and \texttt{fact\_bias}).

\begin{figure}[!t]
\centering
\includegraphics[width=\columnwidth]{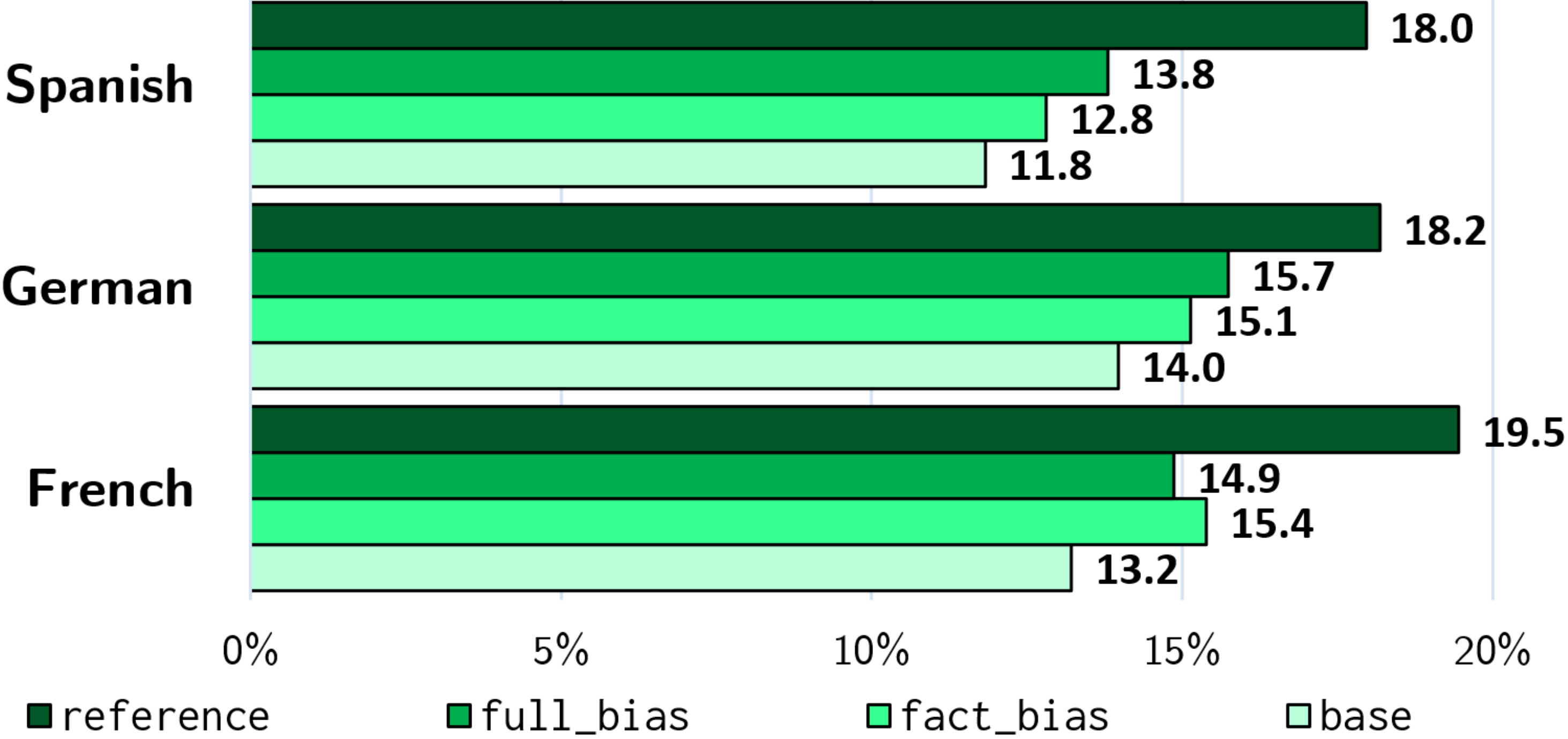}
\caption{\label{fig:cbong_speaker}Speaker classification accuracy of our continuous bag-of-n-grams model.}
\end{figure}

The results are reported in Figure \ref{fig:cbong_speaker}.
As can be seen from the figure, it is easier to predict the author of a sentence from the output of speaker-specific models than from the baseline.
This demonstrates that explicitly incorporating information about the author of a sentence allows for better transfer of personal traits during translations, although the difference from the ground truth demonstrates that this problem is still far from solved.
Appendix \ref{sec:qual_examples} shows qualitative examples of our model improving over the baseline.

\subsection{Further experiments on the Europarl corpus}

One of the quirks of the TED talks is that the speaker annotation correlates with the topic of their talk to a high degree. Although the topics that a speaker talks about can be considered as a manifestation of speaker traits, we also perform a control experiment on a different dataset to verify that our model is indeed learning more than just topical information. Specifically, we train our models on a speaker annotated version of the Europarl corpus \cite{rabinovich-EtAl:2017:EACLlong}, on the \texttt{en-de} language pair\footnote{available here: \url{https://www.kaggle.com/ellarabi/europarl-annotated-for-speaker-gender-and-age/version/1}}.

We use roughly the same training procedure as the one described in \S\ref{sec:exp_known_setup}, with a random train/dev/test split since none is provided in the original dataset. Note that in this case, the number of speakers is much lower (747) whereas the total size of the dataset is bigger ($\approx$300k).

We report the results in table \ref{tab:europarl_results}. Although the difference is less salient than in the case of SATED, our factored bias model still performs significantly better than the baseline ($+0.83$ BLEU). This suggests that even outside the context of TED talks, our proposed method is capable of improvements over a speaker-agnostic model.

\section{Related work}

Domain adaptation techniques for MT often rely on data selection \cite{moore-lewis:2010:Short,li-EtAl:2010:PAPERS3,chen-EtAl:2017:NMT,wang-EtAl:2017:EMNLP20174}, tuning \cite{luong2015stanford,micelibarone-EtAl:2017:EMNLP2017}, or adding domain tags to NMT input \cite{chu-dabre-kurohashi:2017:Short}.
There are also methods that fine-tune parameters of the model on each sentence in the test set \cite{li2016one}, and methods that adapt based on human post-edits \cite{turchi2017continuous}, although these follow our baseline adaptation strategy of tuning all parameters.
There are also partial update methods for transfer learning, albeit for the very different task of transfer between language pairs \cite{zoph-EtAl:2016:EMNLP2016}.

Pioneering work by \newcite{mima1997improving} introduced ways to incorporate information about speaker role, rank, gender, and dialog domain for rule based MT systems.
In the context of data-driven systems, previous work has treated specific traits such as politeness or gender as a ``domain'' in domain adaptation models and applied adaptation techniques such as adding a ``politeness tag'' to moderate politeness \cite{sennrich-haddow-birch:2016:N16-1}, or doing data selection to create gender-specific corpora for training \cite{rabinovich-EtAl:2017:EACLlong}.
The aforementioned methods differ from ours in that they require explicit signal (gender, politeness\ldots) for which labeling (manual or automatic) is needed, and also handle a limited number of ``domains'' ($\approx2$), where our method only requires annotation of the speaker, and must scale to a much larger number of ``domains'' ($\approx1,800$).

\begin{table}[tb]
\centering
\begin{tabular}{lc} & en-de \\ \hline
\texttt{base}  & 26.04\\ \hline
\texttt{spk\_token}  & 26.49\\ \hline
\texttt{full\_bias}  & 26.44\\ \hline
\texttt{fact\_bias} & \textbf{26.87}\\ \hline
\end{tabular}
\caption{\label{tab:europarl_results}Test BLEU on the Europarl corpus. Scores significantly ($p<0.05$) better than the baseline are written in bold}
\end{table}

\section{Conclusion}

In this paper, we have explained and motivated the challenge of modeling the speaker explicitly in NMT systems, then proposed two models to do so in a parameter-efficient way. We cast this problem as an extreme form of domain adaptation and showed that, even when adapting a small proportion of parameters (the softmax bias, $<0.1\%$ of all parameters), allowed the model to better reflect personal linguistic variations through translation.

We further showed that the number of parameters specific to any person could be reduced to as low as 10 while still retaining better scores than a baseline for some language pairs, making it viable in a real world application with potentially millions of different users.

\section*{Acknowledgements}

The authors give their thanks the anonymous reviewers for their useful feedback which helped make this paper what it is, as well as the members of Neulab who helped proof read this paper and provided constructive criticism.
This work was supported by a Google Faculty Research Award 2016 on Machine Translation.

\bibliography{acl}
\bibliographystyle{acl_natbib}

\clearpage
\appendix

\section{Detailed model description}
\label{sec:model_details}

\paragraph{Word embeddings}

We embed source and target words in a low dimensional space with embedding matrices $E_S\in\mathbb{R}^{\vert V_S\vert\times d_{\text{emb}}}$, $E_T\in\mathbb{R}^{\vert V_T\vert\times d_{\text{emb}}}$.
Each word vector is initialized at random from $\mathcal{N}(0,\frac{1}{\sqrt{d_{\text{emb}}}})$. We use $d_{\text{emb}} = 512$.

\paragraph{Encoder}

Our encoder is a one layer bidirectional LSTM with dimension $d_h=512$.
For a source sentence $e = e_1, \dots, e_{|e|}$ the concatenated output of the encoder is thus of shape $\vert e\vert\times 2d_h$.

\paragraph{Attention}

We use a multilayer perceptron attention mechanism: given a query $h_t$ at step $t$ of decoding and encodings $x_1,\ldots,x_{\vert e\vert}$, the context vector $c_t$ is computed according to:
\begin{equation}
\begin{split}
    \alpha_{it} &=V_a^T\tanh(W_{a}e_i+W_{ah}h_t+b_a)\\
    c_t&=\sum_i\alpha_{it}x_i,\\
\end{split}
\end{equation}
where $V_a, W_a, W_{ah}, b_a$ are learned parameters. We choose $d_a=256$ as the dimension of the intermediate layer.

\paragraph{Decoder}

The decoder is a single layer LSTM of dimension $d_h=512$. At each timestep $t$, it takes as input the previous word embedding $w_{t-1}$ and the previous context $c_{t-1}$. Its output $h_t$ is used to compute the next context vector $c_t$ and the distribution over the next possible target words $w_t$:
\begin{equation} \label{eq:output_dist}
\begin{split}
    o_t &=W_{oh}h_t + W_{oc}c_t + W_{ow} E_{Tw_{t-1}} + b_o\\
    p_t&=\text{softmax}(E_To_t + b_T),\\
\end{split}
\end{equation}
where $W_{o*}, b_o, b_T$ are learned parameters, $E_T$ is the target word embedding matrix and $ E_{Tw_{t-1}}$ is the embedding of the previous target word.

\paragraph{Learning paradigm}
We employ several techniques to improve training. First, we are using the same parameters for the target word embeddings and the weights of the softmax matrix \cite{press-wolf:2017:EACLshort}. This reduces the number of total parameters and in practice this gave slightly better BLEU scores.

We apply dropout \cite{srivastava2014dropout} between the output layer and the softmax layer, as well as within the LSTM (using the variant presented in \newcite{gal2016theoretically}). We also drop words in the target sentence with probability 0.1 according to \newcite{iyyer-EtAl:2015:ACL-IJCNLP}. Intuitively, this forces the decoder to use the conditional information.

In addition to this, we implement label smoothing as proposed in \newcite{szegedy2016rethinking} with a smoothing coefficient 0.1. We noticed improvements of up to 1 BLEU point with this additional regularization term.

\section{Training process}
\label{sec:training}

We first train each model using the Adam optimizer \cite{kingma2014adam} with learning rate 0.001 (we clip the gradient norm to 1). The data is split into batches of size 32 where every source sentence has the same length.
We evaluate the validation perplexity after each epoch. Whenever the perplexity doesn't improve, we restart the optimizer with a smaller learning rate from the previous best model \cite{denkowski2017stronger}.
Training is stopped when the perplexity doesn't go down for 3 epochs. We then perform a tuning step: we restart training with the same hyper-parameters except for using simple stochastic gradient descent and gradient clipping at a norm of 0.1, which improved the validation BLEU by 0.3-0.9 points.

\section{User classifier}
\label{sec:classifier}


In our analysis, we use a classifier to estimate which user wrote each output, which we describe more in this section.

The model uses a continuous bag of $n$-grams where $v_{\text{n-gram}}$ is a parameter vector for a paticular n-gram and the probability of speaker $s$ for sentence $f$ is given by:
\begin{equation}
\begin{split}
p(s\mid f)&\propto w_s^Th_f+b_s\\
h_f&=\frac{1}{\#\{\text{n-gram}\in f\}}(\sum_{\text{n-gram}\in f}v_{\text{n-gram}})\\
\end{split}
\end{equation}

The size of hidden vectors is 128.
We limit n-grams to unigrams and bigrams.
We estimate the parameters with Adam and a batch size of 32 for 50 epochs.

\section{Qualitative examples}
\label{sec:qual_examples}

Table \ref{tab:known_qual} shows examples where our \texttt{full\_bias}/\texttt{fact\_bias} model helped translation by favoring certain words as opposed to the baseline in \texttt{en-fr}.

\begin{table*}[!ht]
\centering
\begin{tabular}{ll}
\hline\hline
Talk                       & Andrew McAfee : What will future jobs look like? \\\hline
Source                     & but the middle class is clearly under huge threat right now . \\ 
Reference                  & mais la classe moyenne fait aujourd' hui face \`{a} une grande menace . \\
\texttt{base}              & mais la classe moyenne est clairement une menace \'{e}norme en ce moment . \\
\texttt{full\_bias} & mais la classe moyenne est clairement maintenant \textcolor{red}{dans} une grande menace . \\
\texttt{fact\_bias} & mais la classe moyenne est clairement \textcolor{red}{en} grande menace en ce moment . \\
\hline\hline
Talk                       & Olafur Eliasson : Playing with space and light \\\hline
Source                     & the show was , in a sense , about that . \\ 
Reference                  & le spectacle \'{e}tait , dans un sens , \`{a} propos de cela . \\
\texttt{base}              & le spectacle \'{e}tait , \textcolor{red}{en} un sens , \`{a} propos de \textcolor{red}{\c{c}a} . \\
\texttt{full\_bias} & le spectacle \'{e}tait , \textcolor{red}{dans} un sens , \`{a} propos de \textcolor{red}{cela} . \\
\texttt{fact\_bias} & le spectacle \'{e}tait , \textcolor{red}{dans} un sens , \textcolor{red}{\`{a} ce sujet} . \\
\hline\hline
\multirow{2}{*}{Talk}                       & Lona Szabo de Carvalho : 4 lessons I learned from taking a stand\\&against drugs and gun violence \\\hline
Source                     & we need to make illegal drugs legal . \\ 
Reference                  & nous avons besoin de rendre les drogues ill\'{e}gales , l\'{e}gales . \\
\texttt{base}              & nous devons faire des \textcolor{red}{m\'{e}dicaments} ill\'{e}gaux . \\
\texttt{full\_bias} & nous devons faire des \textcolor{red}{drogues} ill\'{e}gales . \\
\texttt{fact\_bias} & nous devons produire des \textcolor{red}{drogues} ill\'{e}gales . \\
\hline\hline
Talk                       & Wade Davis: On the worldwide web of belief and ritual\\\hline
Source                     & a people for whom blood on ice is not a sign of death , but an affirmation of life . \\ 
\multirow{2}{*}{Reference}                  & un peuple pour qui du sang sur la glace n' est pas un signe de mort\\& mais une affirmation de la vie . \\
\multirow{2}{*}{\texttt{base}}              & \textcolor{red}{une personne} pour qui \textcolor{red}{le} sang sur la glace n' est pas un signe de mort ,\\& mais une affirmation de \textcolor{red}{la} vie . \\
\multirow{2}{*}{\texttt{full\_bias}} & \textcolor{red}{un peuple} pour qui \textcolor{red}{le} sang sur la glace n' est pas un signe de mort , \\&mais une affirmation de \textcolor{red}{la} vie . \\
\multirow{2}{*}{\texttt{fact\_bias}} & \textcolor{red}{un peuple} pour qui sang sur la glace n' est pas un signe de mort , \\&mais une affirmation de vie . \\
\hline

\end{tabular}
\caption{\label{tab:known_qual}Examples where our proposed method helped improve translation.}
\end{table*}

\end{document}